\documentclass[sn-mathphys-num]{sn-jnl}

\usepackage{graphicx}%
\usepackage{multirow}%
\usepackage{array}
\usepackage{amsmath,amssymb,amsfonts}%
\usepackage{amsthm}%
\usepackage{mathrsfs}%
\usepackage[title]{appendix}%
\usepackage{tabularx}
\usepackage{xcolor}%
\usepackage{textcomp}%
\usepackage{manyfoot}%
\usepackage{booktabs}%
\usepackage{algorithm}%
\usepackage{algorithmicx}%
\usepackage{algpseudocode}%
\usepackage{listings}%

\theoremstyle{thmstyleone}%
\theoremstyle{thmstyletwo}%

\theoremstyle{thmstylethree}%

\begin{document}

\title[Article Title]{Trajectory Prediction in Dynamic Object Tracking: A Critical Study}


\author*[1]{\fnm{Zhongping} \sur{Dong}}\email{zhongping.dong@ucdconnect.ie}
\equalcont{These authors contributed equally to this work.}

\author[2]{\fnm{Liming} \sur{Chen}}\email{limigchen0922@dlut.edu.cn}
\equalcont{These authors contributed equally to this work.}

\author[1]{\fnm{Mohand Tahar} \sur{Kechadi}}\email{tahar.kechadi@ucd.ie}
\equalcont{These authors contributed equally to this work.}

\affil*[1]{\orgdiv{School of Computer Science}, \orgname{University College Dublin}, \orgaddress{\street{Belfield}, \city{Dublin 4}, \postcode{D04 V1W8}, \state{Dublin}, \country{Ireland}}}

\affil[2]{\orgdiv{School of Computer Science and Technology}, \orgname{Dalian University of Technology}, \orgaddress{\street{No.2 Linggong Road}, \city{Dalian}, \postcode{116024}, \state{Liaoning}, \country{China}}}



\abstract{This study provides a detailed analysis of current advancements in dynamic object tracking (DOT) and trajectory prediction (TP) methodologies, including their applications and challenges. It covers various approaches, such as feature-based, segmentation-based, estimation-based, and learning-based methods, evaluating their effectiveness, deployment, and limitations in real-world scenarios. The study highlights the significant impact of these technologies in automotive and autonomous vehicles, surveillance and security, healthcare, and industrial automation, contributing to safety and efficiency. Despite the progress, challenges such as improved generalization, computational efficiency, reduced data dependency, and ethical considerations still exist. The study suggests future research directions to address these challenges, emphasizing the importance of multimodal data integration, semantic information fusion, and developing context-aware systems, along with ethical and privacy-preserving frameworks.}




\maketitle

\section{Introduction}\label{sec1}

In the domain of computer vision and automation, the integration of Dynamic Object Tracking (DOT) and Trajectory Prediction (TP) is crucial for developing intelligent systems capable of real-time environmental interaction and understanding \cite{zhong2023rspt}. This integration, essential for autonomous navigation, robotics, and traffic management, leverages machine learning and sensor technology advancements to track moving objects precisely and efficiently \cite{bharilya2024machine}. DOT focuses on monitoring objects in motion \cite{yilmaz2006object}, whereas TP predicts their future movements \cite{wang2022stepwise} to enhance system autonomy and adaptability. Employing cutting-edge computer vision and machine learning techniques will facilitate accurate trajectory prediction and proactive decision-making, significantly advancing more responsive and intelligent technological ecosystems \cite{cao2022dynamic}.

\begin{figure}[h]
\centering
\includegraphics[width=0.9\textwidth]{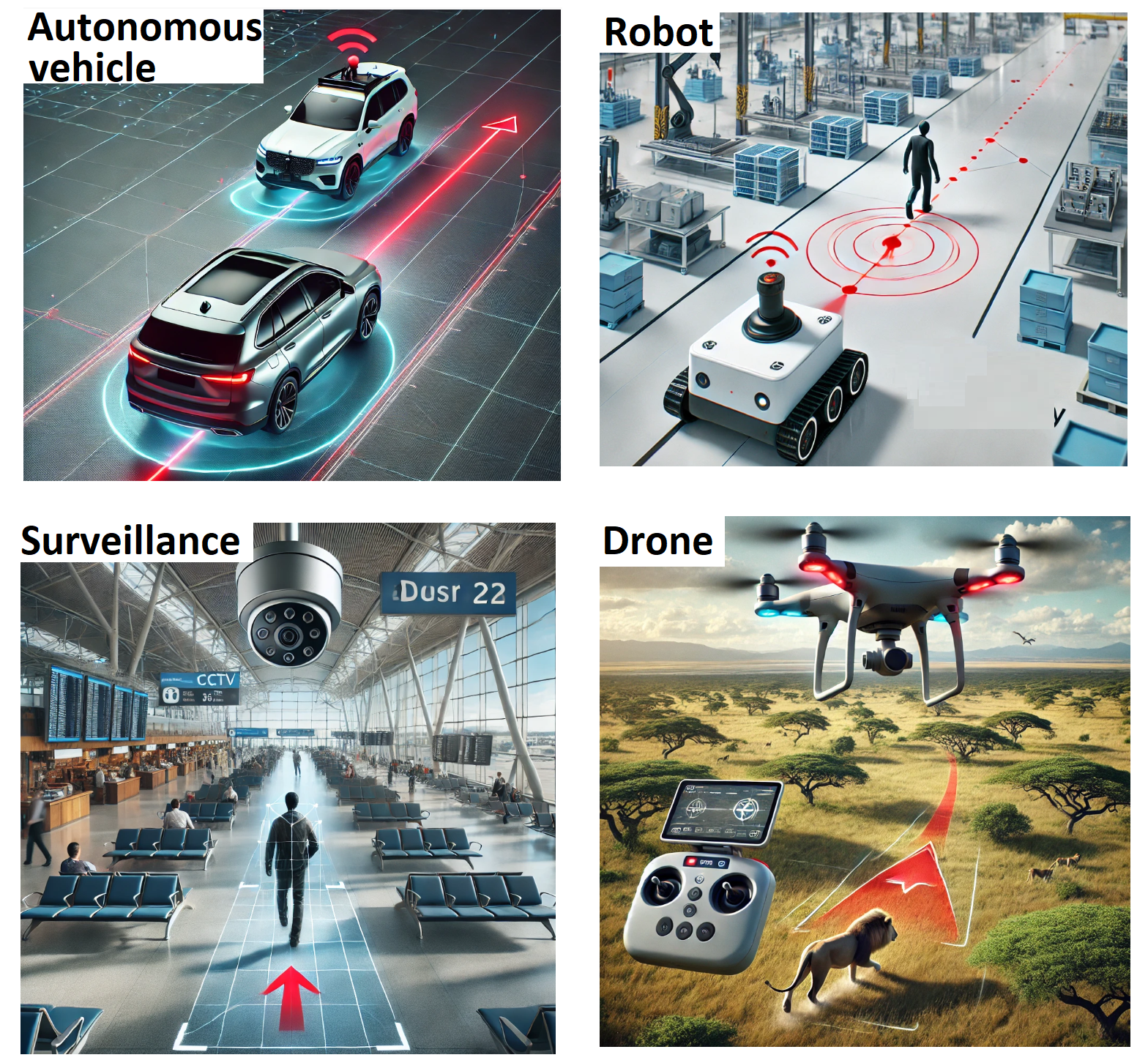}
\caption{Integration of Dynamic Object Tracking and Trajectory Prediction: Extensive Applications Across Multiple Industries.}\label{fig: DOT-TP Application}
\end{figure}

Dynamic Object Tracking and Trajectory Prediction are critically interdependent and have extensive applications across multiple industries, augmenting both efficiency and safety \cite{rudenko2020human}, as shown in Figure \ref{fig: DOT-TP Application}. These technologies are fundamental in autonomous vehicles for real-time detection and response \cite{biermann2019football}. In industrial contexts, robots in automated warehouses, such as those employed by Amazon, utilize these systems for effective navigation and task execution \cite{ng2020adaptive}. In security and surveillance domains, tracking is pivotal for monitoring activities, as evidenced in airport security measures \cite{ali2019traffic}.  The versatility of DOT and TP extends into the sports and entertainment industries, where technologies like RFID are employed by the NFL’s Next Gen Stats to provide in-depth player movement analytics, thus enriching fan experiences and enhancing strategic game development \cite{biermann2019football}. In healthcare, these technologies aid in monitoring elderly care and significantly reducing fall incidents \cite{adeli2021tripod}. Drones also benefit from improved operational efficiency through the adoption of these technologies \cite{conte2021drone}. Additionally, in urban traffic management, DOT and TP contribute to optimizing traffic flow and enhancing safety \cite{yang2023improved}. These multifaceted applications stand as testaments to the broad and versatile applicability of these technologies in contemporary society.

\subsection{Challenges}
\label{sec:Challenges}
DOT and TP face significant challenges due to the unpredictability and complexity of real-world environments. Occlusions, visibility limitations in densely populated regions, and erratic movements of entities, like pedestrians, are major obstacles to consistent tracking and accurate trajectory prediction \cite{wu2020visual}. External conditions and sensing technology constraints further complicate tracking, especially in outdoor environments \cite{rudenko2020human}. These challenges are outlined in Table \ref{tab: Challenges}. This table draws on benchmark datasets like MOTChallenge\cite{dendorfer2021motchallenge} and KITTI \cite{geiger2013vision}, supplemented by illustrative imagery, to provide deep insights into the encountered complexities. It showcases the various scenarios that introduce these challenges and emphasizes the critical need for specialized, adaptive methods. 

\begin{table}[h]
\centering
\caption{Overview of Challenges in Dynamic Object Tracking and Trajectory Prediction}
\label{tab: Challenges}

\begin{tabular}{|>{\raggedright\arraybackslash}p{2.59cm}|>{\raggedright\arraybackslash}p{4.1cm}|>{\raggedright\arraybackslash}p{2.6cm}|}

\hline

{\bf Challenge} & {\bf Description} & {\bf Example}  \\
\hline
Varied Object Speeds & Objects being tracked can vary greatly in moving at different speeds. & \raisebox{\dimexpr \ht\strutbox-\height}{\includegraphics[width=1in, keepaspectratio]{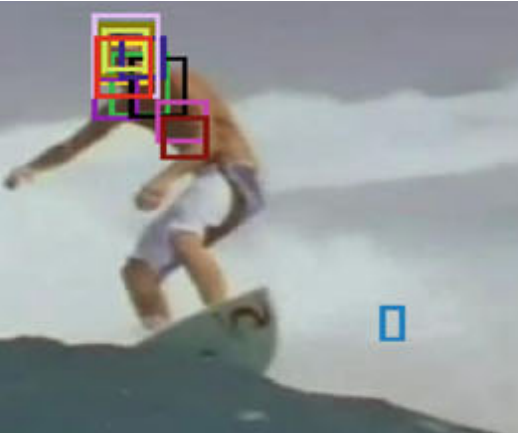}}\\
\hline

Occlusion & Objects of interest are often partially or fully obscured by other objects, complicating their tracking and trajectory prediction. &\raisebox{\dimexpr \ht\strutbox-\height}{\includegraphics[width=1in, keepaspectratio]{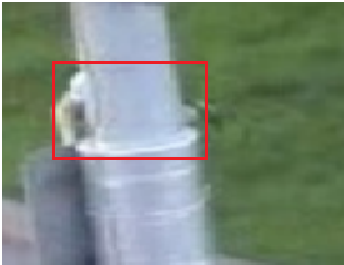}}\ \\
\hline

Perspective and Scale Changes & The perspective and scale of objects change as they move relative to the camera, affecting tracking accuracy. & \raisebox{\dimexpr \ht\strutbox-\height}{\includegraphics[width=1in, keepaspectratio]{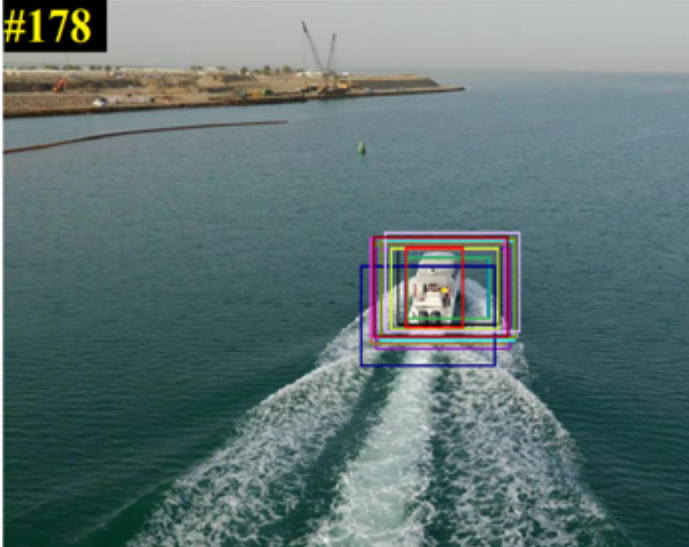}}  \\
\hline

Background Clutters & The background near the target has a similar color or texture as the target & \raisebox{\dimexpr \ht\strutbox-\height}{\includegraphics[width=1in, keepaspectratio]{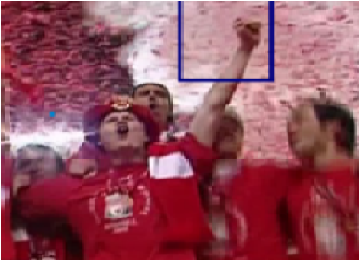}} \\
\hline

Lighting and Weather Conditions & Variations in lighting and weather can significantly affect the appearance and visibility of objects. & \raisebox{\dimexpr \ht\strutbox-\height}{\includegraphics[width=1in, keepaspectratio]{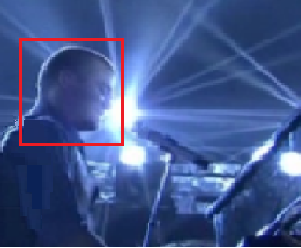}}  \\
\hline

Non-linear Trajectories & Objects may move in unpredictable patterns, making it difficult to accurately predict future positions. & \raisebox{\dimexpr \ht\strutbox-\height}{\includegraphics[width=1in, keepaspectratio]{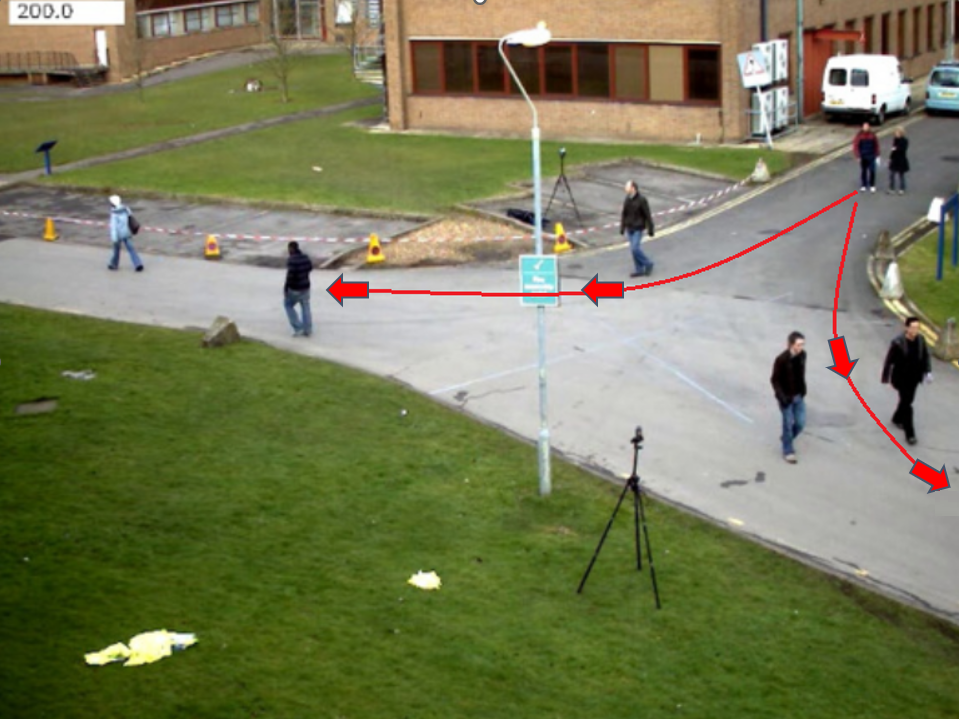}} \\
\hline

Multi-object Tracking & Scenes often contain multiple objects moving simultaneously, complicating tracking and prediction. & \raisebox{\dimexpr \ht\strutbox-\height}{\includegraphics[width=1in, keepaspectratio]{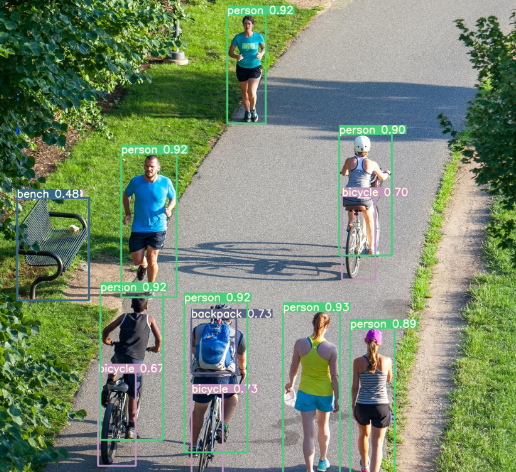}}\\
\hline

\end{tabular}
\end{table}

\subsection{Study Relevance}
\label{sec:Reviews}
This study addresses a crucial gap in the DOT field, which is the integration of trajectory prediction with dynamic tracking techniques. DOT (SOT and MOT) and TP have been viewed as separate domains, as shown in the literature devoted to DOT \cite{balaji2017survey,fiaz2018tracking, bi2019review,han2022single,fu2023siamese} or TP \cite{georgiou2018moving,rudenko2020human,sighencea2021review,huang2022survey,golchoubian2023pedestrian,schuetz2023review}. This allowed the development of advanced techniques in each independently of the other.
Similarly, TP studies tend to be constrained by a limited perspective, focusing exclusively on specific user groups, thus neglecting its wider applicability.

SOT focuses on continuously observing a singular, predefined object within a video stream. SOT's primary objective is to accurately observe the position of this object from frame to frame throughout a video sequence. This process begins with the identification or initial selection of the target object in the first frame of the video. Subsequent frames require the system to not only detect the presence of this object despite potential changes in scale, orientation, or lighting but also to determine its movement and update its location accordingly. MOT focuses on tracking multiple objects and their interactions, employing data association and tracking-by-detection techniques to manage multiple identities and locations simultaneously. The challenges include handling occlusions where the object may be temporarily obscured, changes in appearance, and maintaining tracking reliability over diverse environmental conditions. 

Upon a comprehensive review of the available literature on dynamic object tracking, we have identified a triad of significant gaps in the scholarly discourse:
\begin{enumerate}
    \item {\bf Lack of Synergies between DOT and TP domains}: DOT and TP have often been treated as distinct domains, overlooking the potential integrative benefits of these topics. This critical study examines how integrating DOT  with TP within the pattern analysis framework could significantly improve the accuracy and robustness of tracking systems.
 
    \item {\bf Lack of Specificity in SOT and MOT Challenges}: Current research inadequately differentiates between SOT and MOT within dynamic environments, a critical oversight for applications demanding precise trajectory predictions such as autonomous driving, surveillance, and robotics. This study aims to thoroughly delineate and explore the unique challenges posed by SOT and MOT. 
    
    \item {\bf Overlooking DOT Key Characteristics}: Essential aspects, such as context awareness, multi-modal approaches, semantic understanding, and privacy-preserving analysis techniques, are frequently disregarded. This results in the development of advanced yet ultimately insufficient solutions that do not fully tackle the environmental complexity (adaptability,  privacy, scalability, ...).
\end{enumerate}

\subsection{Aim and Contributions}
\label{sec:Contributions}
In this study we explore the potential synergies between dynamic object tracking (DOT) and trajectory prediction (TP) to improve the accuracy and robustness of tracking systems. It sets specific objectives: assessing the current state of DOT and TP integration, identifying critical characteristics such as context-awareness and privacy preservation in DOT systems, evaluation methodologies, models, datasets, and metrics. This effort aims to provide a detailed understanding of the challenges and essential features of effective DOT systems.  

Moreover, this study contributes to DOT's theoretical understanding and its practical applications. We address the distinct challenges of SOT and MOT. For instance, we clearly identified their differences, as summarized in Table \ref{tab:Differences Between SOT and MOT}. This comparison highlights the challenges and objectives specific to each tracking type. We introduced a unified model between SOT and MOT on the one hand and TP on the other hand. This model improves system robustness and tracking accuracy.

\section{Problem Statement \& Concepts}
\label{sec:key Definitions}
DOT focuses on the instantaneous identification and monitoring of objects across different frames or environments, providing a detailed insight into their current condition \cite{elhoseny2020multi}. TP extends this understanding by predicting future trajectories using analyses of past and present movements \cite{cong2023dacr}. These two concepts are closely related, and building a tracking system that combines both techniques, thus taking advantage of both, is more suitable for real-time tracking. Their integration creates a dynamic feedback loop whereby DOT's tracking data enhances TP's predictive accuracy, which in turn refines the tracking algorithms, as illustrated in Figure \ref{fig: DOT/TP_feedback_loop}. The system consists of four steps:

\begin{figure}[h]
\centering
\includegraphics[width=0.9\textwidth]{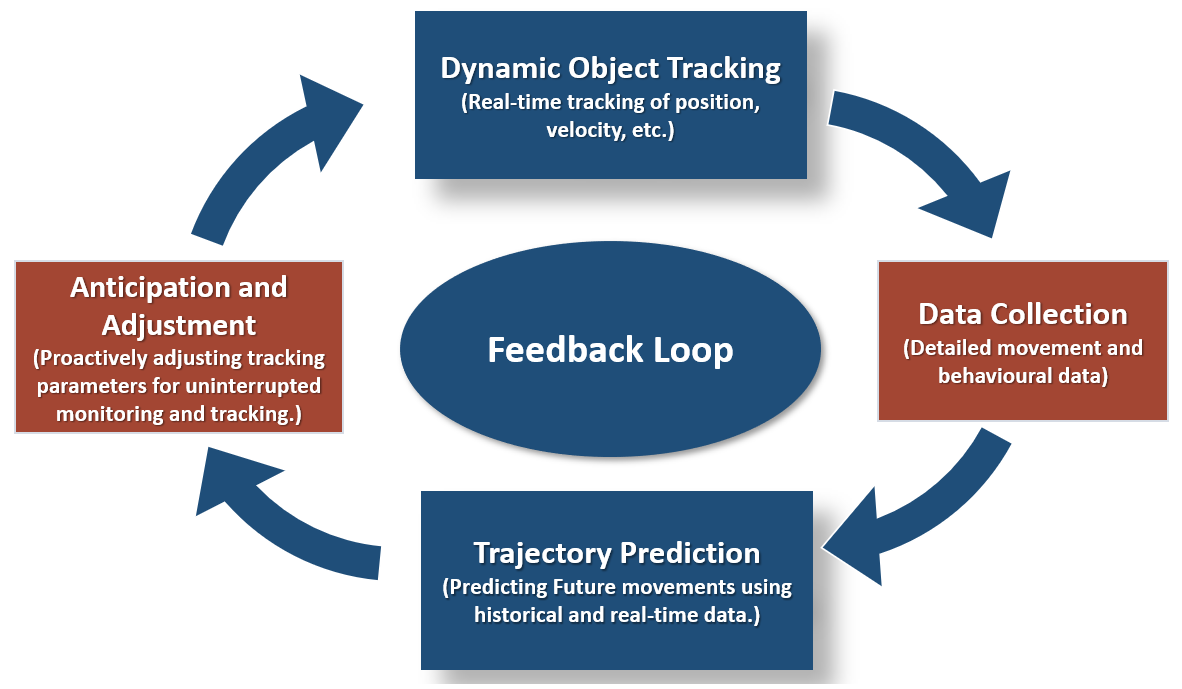}
\caption{The feedback loop between DOT and TP.}
\label{fig: DOT/TP_feedback_loop}
\end{figure}

\begin{enumerate}
  \item The system starts by tracking an object, focusing on its position, velocity, and other relevant parameters to gain some insights about its immediate status.
  \item It collects data about that object, capturing every aspect of its trajectory and emerging patterns over time.
  \item The gathered data is fed into TP techniques to forecast subsequent movements, considering various influencing factors.
  \item TP anticipates the object's next moves. This is critical when the object moves out of sight. By foreseeing potential future movements, the system can proactively adjust tracking parameters, ensuring uninterrupted and accurate monitoring.
\end{enumerate}

The proposed system not only significantly reduces the computational time of both DOT and TP techniques but also improves the accuracy and reliability of prediction and real-time object tracking in complex scenarios. The collaborative use of DOT and TP allows for more efficient capabilities. It exhibits high potential for developing adaptive, responsive, and intelligent predictive tracking systems operating with high precision in dynamic environments. In the following section, we will discuss the system characteristics and the technologies that have been developed so far.

\section{System Characteristics}
\label{sec:SCa}
\subsection{Single and Multiple Object Tracking}
\label{sec:SOT and MOT}

We looked at the differences between SOT and MOT. The results are summarized in Table \ref{tab:Differences Between SOT and MOT}. This comparison highlights the challenges and objectives specific to each tracking type. 
Moreover, the differences between SOT and MOT regarding their trajectory prediction capabilities are summarized in Table \ref{tab:Differences Between SOT and MOT in TP}.

SOT can detect the path of a single object, which simplifies data processing and allows for the use of detailed, object-specific prediction models, which typically results in higher accuracy and lower resource use due to fewer variables. Conversely, MOT may be able to determine the trajectories of multiple objects simultaneously, which is very challenging due to inter-object interactions, occlusions, etc. This complexity necessitates generalized, scalable models capable of handling data from various sources, increasing computational demands and potentially reducing prediction accuracy due to the intricacies of managing multiple dynamic objects. Understanding these fundamental differences is crucial for developing efficient and precise tracking systems tailored to specific applications, whether they require monitoring a single entity or multiple entities within a dynamic environment.

\begin{table}[h]
\caption{Differences Between SOT and MOT}
\label{tab:Differences Between SOT and MOT}
\renewcommand{\arraystretch}{1.5}
\centering

\begin{tabularx}{0.8\columnwidth}{@{} >{\raggedright\arraybackslash}Xp{3.8cm} >{\raggedright\arraybackslash}p{3.8cm} @{}}

\toprule
& {\bf SOT} & {\bf MOT} \\ 

\midrule
{\bf Objective} & Track a single object across a sequence of frames & Track multiple objects simultaneously \\ \hline
{\bf Challenges} & Occlusions, appearance changes & Occlusions, identity switches, interactions \\ \hline
{\bf Methodologies} & Template matching, discriminative models & Data association, tracking-by-detection \\ \hline
{\bf Applications} & Autonomous navigation, sports analytics & Surveillance, crowd monitoring \\ \hline
{\bf Precision} & Generally higher due to focus on a single target & Potentially lower due to complexities of tracking multiple objects \\ \hline
{\bf Computational Complexity} & Lower compared to MOT & Higher due to the need to handle multiple objects \\ 

\bottomrule
\end{tabularx}
\end{table}

\begin{table}[h]
\caption{Differences Between SOT and MOT in Trajectory Prediction}
\label{tab:Differences Between SOT and MOT in TP}
\renewcommand{\arraystretch}{1.5}
\centering

\begin{tabularx}{0.8\columnwidth}{@{} >{\raggedright\arraybackslash}Xp{3.6cm} >{\raggedright\arraybackslash}p{4.0cm} @{}}

\toprule
& {\bf SOT for Trajectory Prediction} & {\bf MOT for Trajectory Prediction}  \\ 
\midrule

{\bf Objective} & Predict the trajectory of a single object & Predict trajectories of multiple objects simultaneously \\ \hline
{\bf Challenges} & Maintaining accuracy in dynamic environments & Handling interactions and occlusions among multiple objects \\ \hline
{\bf Data Processing} & Simpler, as data pertains to one object only & More complex due to data from multiple sources and interactions \\ \hline
{\bf Prediction Models} & Detailed, object-specific models & Generalized, scalable models for multiple objects \\ \hline
{\bf Accuracy} & Generally higher due to the singular focus & May vary, often lower due to complexity of scene and interactions \\ \hline
{\bf Resource Intensity} & Lower, with fewer variables to consider & Higher, due to the need to process and predict multiple trajectories simultaneously \\

\bottomrule
\end{tabularx}
\end{table}

\subsection{DOT Systems and Multi-modality}     
\label{sec:DOTMM}

\begin{sidewaystable}
\centering
\caption{Overview of the Key Features of the DOT and TP. MM: MULTI-MODAL. CA: CONTEXT-AWARE. SU: SEMANTIC UNDERSTANDING. PP: PRIVACY PRESERVING.}
\label{tab:key_features}
\renewcommand{\arraystretch}{1.1}

\begin{tabularx}{\columnwidth}{@{} >{\raggedright\arraybackslash}p{1cm}>
{\raggedright\arraybackslash}p{2.8cm}>
{\raggedright\arraybackslash}p{1.95cm} > 
{\raggedright\arraybackslash}p{1.95cm} >
{\raggedright\arraybackslash}p{6.39cm} >{\centering\arraybackslash}p{0.45cm} >{\centering\arraybackslash}p{0.45cm} >{\centering\arraybackslash}p{0.45cm} >{\centering\arraybackslash}p{0.45cm}@{}}

\toprule
{\bf Paper} &  {\bf Tracking Technology} & {\bf Dataset} & {\bf Metrics} &  {\bf Results} & {\bf MM} & {\bf CA} & {\bf SU} & {\bf PP} \\
\midrule

\cite{wang2021dynamic} & Siamese Network based THOR tracker & TLP, DTB, GOT-10K, \ldots & PR, SR, TPR, TNR, \ldots & Comparable performance on various benchmarks & $\checkmark$ & $\times$ & $\times$ & $\times$ \\

\cite{zhang2021map} & Map-Adaptive Goal-Based Trajectory Prediction & nuScenes & Min1ADE, Min1FDE, et al. & Outperforms other methods, showing minimal performance regression when evaluated in different cities.  &$\checkmark$ & $\times$ & $\times$ &  $\times$ \\

\cite{mehmood2020context} & Correlation Filter and Kalman filter & Color(TC)-128, OTB2013,  \ldots & CLE, DPR & Improved robustness and accuracy in occlusion and challenges.  & $\times$ & \checkmark &  $\times$ & $\times$ \\

\cite{wang2020hierarchical} & Hierarchical Spatiotemporal Context-Aware Correlation Filters & OTB-2013, OTB-2015, VOT-2016, UAV-20L & AUC, OSR, LPR, EAO & Superior tracking performance; real-time tracking; effective in handling occlusion and object deformations. & $\times$ & \checkmark &  $\times$ & $\times$ \\

\cite{shen2022context}& Deep Learning: Conv & VOT2018, OTB100 & EAO, Rob, ACC & Achieves EAO of 0.514 on VOT2018 at 40FPS. & $\times$ & \checkmark & $\times$ & $\times$ \\

\cite{zhang2020spatial} & Spatial Semantic Convolutional Features & OTB-2013 & PSR, Skewness & Improved tracking accuracy by fusing spatial and semantic features; robust against scale variations and occlusions.  & $\times$ & $\times$ & $\checkmark$ & $\times$ \\

\cite{sun2020see} & Semantic Segmentation Network for Ego-Vehicle Trajectory & KITTI & Accuracy, IoU & Improves autonomous driving safety through precise trajectory prediction, minimizing false positives across multiple prediction horizons.  & $\times$ & $\times$ & $\checkmark$ & $\times$ \\

\midrule

\cite{morton2011evaluation} & Updated Mesh Segmentation Algorithm & 3D LIDAR Lawn data & Fragments number, RMS & Achieved 95\% Correct Data Associations & $\checkmark$ & $\times$ & $\times$ & $\times$ \\

\cite{zhou2019semantic} & SegNet, FCN-8s, DilatedResNet and Mask R-CNN & ADE20k Dataset & Pixel ACC., Mean IoU, PSPNet, mAP & A reasonably large batch size is crucial for semantic segmentation. & $\times$ & $\times$ & $\checkmark$ & $\times$ \\

\cite{tian2021robust} & Privacy-Preserving Motion Detection in Encrypted Video & FaceScrub, FW dataset & Detection accuracy, robustness & Achieves robust performance in dynamic surveillance, ensuring privacy without requiring video decryption. & $\times$ & $\times$ & $\times$ &  $\checkmark$ \\

\cite{giorgi2022privacy} & ResNet-50, Autoencoders and YOLOv4 & UCF-CRIME & MSE, PSNR, SSIM, FL-AUC, VL-F1 & Identify video anomalies correctly with a consistent privacy gain. & $\times$ & $\times$ & $\times$ &  $\checkmark$ \\

\cite{ohno2024privacy} & Path image inpainting and 3D point cloud features & Campus LiDARs dataset & F-measure & High privacy-preserving accuracy in pedestrian tracking with a 0.98 F-measure, showcasing robustness in complex environments while maintaining tracking quality.& $\times$ & $\times$ & $\times$ &  $\checkmark$ \\
\bottomrule

\end{tabularx}
\end{sidewaystable}

Multi-modal object tracking involves the integration and fusion of data from multiple sources, such as cameras, radar, and LiDAR, to form a comprehensive understanding of the scene.

For instance, MMOT plays a vital role in urban traffic management by continuously tracking different entities such as pedestrians, bicycles, and vehicles. Each entity requires unique handling capabilities, such as adaptation to speed variations and intricate maneuvering in densely populated areas. The integration of these models ensures robust and accurate tracking by dynamically adjusting to real-time environmental inputs and object behaviors. This adaptability is particularly effective in enhancing traffic flow management systems and improving overall safety and efficiency in urban settings.

Recent research, \cite{kumar2020novel, wang2023trajectory}, highlights the trend towards leveraging diverse arrays of data types to refine tracking precision. They pioneered a multi-cue feature fusion method that integrates Local Binary Patterns with Histograms of Oriented Gradients to illustrate the advantages of using various feature types. However, this approach faces challenges in computational efficiency, scalability, and adaptability across different environments and targets, compounded by a dependency on the quality of input data, particularly in adverse conditions. In contrast, \cite{vincent2020dynamic} demonstrated the use of a combination of Deep Neural Networks (DNN), Extended Kalman Filters (EKF), and Visual SLAM to improve localization and loop closure detection in complex datasets like TUM and COCO, illustrating the benefits of integrating multiple modalities to manage complex visual data. This trend is also evident in \cite{kim2008real} and \cite{wang2023trajectory}, which employ Kalman filters and Siamese Networks to achieve real-time performance and high-quality trajectories. Despite these advances, these systems' computational demands and practical implementation remain challenging, underscoring the ongoing need to balance robustness with efficiency in real-world applications.

\subsection{Context-Aware Object Tracking}
\label{sec:Multi-Context Object Tracking}

Context-aware object tracking is crucial in dynamic object tracking (DOT) research, as it integrates environmental and situational data to resolve uncertainties in cluttered or changing environments. By incorporating contextual information, such as scene layout and object interactions, context-aware methods enable systems to predict and adapt to changes, boosting adaptability and accuracy. For example, in autonomous vehicle navigation, context-aware tracking uses data on road layouts, traffic conditions, and pedestrian movements to anticipate object paths and optimize responses to unexpected scenarios, ensuring operational safety and efficiency in complex environments.

Recent studies, \cite{mehmood2020context, wang2020hierarchical}, illustrate the practical enhancements that context-aware systems bring to object tracking. \cite{wang2020hierarchical} utilized Hierarchical Spatiotemporal Context-Aware Correlation Filters to improve tracking performance significantly in scenarios involving occlusions and object deformations. \cite{shen2022context} extended these capabilities by using deep learning techniques, such as a high Effective Average Overlap (EAO) in challenging datasets like VOT2018. However, these systems often face challenges related to computational demand and the limited availability of training samples, which can hinder their effectiveness in online optimization. Furthermore, the current scope of these context-aware methods, often confined to controlled test environments, underscores the need for further development to enhance their adaptability in more dynamic real-world applications such as autonomous driving and urban surveillance. This research highlights the potential and the limitations of context-aware tracking technologies, emphasizing the need for continued innovation and refinement in this field.

\subsection{Semantic Information in DOT}
\label{sec:Semantic Understanding}

Dynamic object tracking (DOT) has evolved to prioritize semantic understanding. By leveraging object attributes and scene context, we've significantly improved object identification and tracking, especially in complex environments \cite{ma2024semantic}. Semantic understanding empowers DOT systems to discern and respond to different object types and their behaviors. For example, in security surveillance, it can differentiate between a suspicious vehicle and one simply parked. By interpreting these semantic cues, systems can prioritize and allocate resources more effectively, focusing on objects of interest. This nuanced understanding is crucial for maintaining situational awareness and operational efficiency in dynamic environments \cite{lo2021dynamic}. It enhances the robustness and responsiveness of tracking systems.

\cite{zhou2019semantic} and \cite{zhang2020spatial} have employed advanced segmentation networks and convolutional features to improve semantic segmentation and tracking accuracy. Utilizing datasets ADE20k and OTB-2013, these studies demonstrate that the integration of semantic information significantly reduces errors and enhances interaction with the environment. This approach is particularly valuable in applications involving autonomous vehicles and augmented reality, where a deep comprehension of both content and context is necessary. The ongoing research in semantic understanding highlights the need for continued development in semantic awareness, recognizing that accurate semantic information is indispensable for improving motion pattern analysis and tracking accuracy, especially in environments characterized by obstacles or frequent scene changes \cite{yang2018unifying}. This focus on enhancing semantic awareness is crucial for advancing the efficacy and adaptability of tracking technologies in increasingly complex and dynamic settings.

\subsection{Privacy-Preserving Techniques}
\label{sec:Privacy-Preserving Techniques}
In the context of escalating privacy concerns in video data processing, privacy-preserving techniques have become crucial in dynamic object tracking (DOT), especially in applications involving personal data or sensitive locations. These methods ensure that tracking activities are conducted without compromising individual privacy, balancing effective surveillance and tracking capabilities with the protection of individual rights. In environments involving individuals, privacy-preserving technologies enhance tracking systems by obscuring personally identifiable information, thus maintaining privacy without sacrificing functionality \cite{moon2024object}. For instance, in retail analytics, such technologies anonymize shoppers' identities while analyzing their movement patterns to optimize store layout and product placements \cite{d2015privacy}. Techniques such as data encryption, anonymization, or obfuscation ensure that the system gathers valuable insights from tracking data without exposing the personal details of the individuals being tracked. This compliance with privacy regulations, such as the GDPR, builds public trust and facilitates the wider acceptance and deployment of advanced tracking technologies in sensitive settings.

In the realm of privacy-preserving technologies, recent studies  highlight the critical need to maintain privacy in tracking applications \cite{tian2021robust} and \cite{ohno2024privacy}. The former developed techniques for detecting motion in encrypted videos without decryption, ensuring user privacy while maintaining robust performance. Meanwhile, the later employed innovative methods like path image inpainting combined with 3D point cloud features to preserve privacy in pedestrian tracking, achieving high accuracy in privacy-sensitive datasets. Furthermore, \cite{giorgi2022privacy} introduced a significant advancement with a framework designed for privacy-conscious video anomaly detection that uses differential privacy principles within a two-stream ResNet-50 architecture. This study represents a key development in harmonizing the objectives of preserving privacy and ensuring accuracy in anomaly detection, providing a model for balancing advanced data processing needs with strict privacy requirements. These efforts are crucial for advancing the integration of privacy-preserving mechanisms in modern, data-driven research, particularly in applications involving multi-modal data fusion and dynamic tracking.

\section{Dynamic Object Tracking Models}
\label{sec:Models for DOT and TP}
There have been a plethora of learning models used for dynamic object tracking in various systems. These models can be classified into different categories: generative, discriminative, Siamese network-based, deep learning-based, graph-based, and data association-based models, each designed to address specific challenges inherent to tracking.

\subsection{Generative Models}
\label{subsec:Generative Models}
For DOT, generative models focus on capturing and learning the appearance of an object from collected data, enabling them to distinguish the target from the background across successive video frames. This capability is especially beneficial for tracking in environments where the visibility of the object may be compromised by occlusion or rapid movements. Predominantly, generative models include techniques such as Template Matching and Appearance Models, each catering to specific challenges in tracking. Template Matching provides simplicity and effectiveness under stable conditions, while Appearance Models offer flexibility and resilience in more complex environments. The selection of a specific model should be guided by the particular tracking requirements and the conditions of the environment. Despite their strengths, it is important to acknowledge that generative models are limited to dealing with rapid movements and occlusions, which may reduce their effectiveness in less controlled settings.

\paragraph{Template Matching} involves comparing a predefined template or a reference image of the object against subsequent frames in a video sequence to identify and track the object's position. This method is praised for its high accuracy and robustness in controlled environments, where the appearance of the object remains consistent throughout the sequence. It excels in scenarios with limited variation in object appearance and minimal occlusion \cite{yan2019adaptive}, \cite{sun2020fast}.

\paragraph{Appearance Models} are engineered to adaptively learn and update an object visual representation, making them indispensable for dynamic tracking. These models continuously refine the object's representation to accommodate changes in viewpoint, lighting, or deformation \cite{zhang2007efficient}. 

\subsection{Discriminative Models}
\label{subsec:Discriminative Models}
Discriminative models focus on differentiating the target object from its background rather than reconstructing the object's appearance. They are particularly valuable in scenarios where the object's visibility is compromised by complex backgrounds or dynamic changes within the scene. Their main strengths lie in their robustness against occlusions and adaptability to appearance changes. The main techniques used in these models include SVMs \cite{bhat2019learning} and Correlation Filters\cite{kart2019object}. However, their performance may degrade in scenarios involving extreme object deformations or environments where the background closely mimics the object.

{\bf SVM Classifiers} are heavily used in discriminative models. The main idea is to isolate the object from the background and identify its precise localization across video frames by performing a binary classification. SVMs and their variants (e.g., integration of fuzzy logic \cite{zhang2015single} and convolutional networks \cite{li2017object}) adapt dynamically, being trained on the fly with data extracted directly from the video to adjust to changes in the object's appearance.

{\bf Correlation Filters} are utilized to accurately model and predict the object's location in video sequences by optimizing the correlation response, therefore enhancing tracking accuracy. Fundamentally, these methods involve the design of an optimal image filter trained on the target enclosed by a bounding box in the initial frame. Subsequent frames crop around the predicted position to maintain focus on the moving target. Despite variations in lighting, scale, and pose, these approaches have maintained high tracking accuracy, demonstrating significant advances in visual tracking technology \cite{bolme2010visual} \cite{ma2022correlation}.

\subsection{Siamese Network-Based Models}
\label{subsec:Siamese Network-Based Models}
Siamese Network-Based Models offer a robust architecture that uses twin networks to compare features across different image patches. Examples of such models include SiamFC\cite{xu2020real} and SiamRPN\cite{fang20203d}, which are particularly effective due to their ability to encode the appearance of an object and assess it against candidate regions in successive video frames. This makes them highly suitable for real-time applications that demand rapid and reliable tracking, such as surveillance and autonomous vehicle navigation. Siamese networks are robust to variations in object appearance and operational efficiency, crucial for maintaining performance in dynamic tracking environments.

\subsection{Deep Learning Models}
\label{subsec:Deep Feature-Based Models}
Deep Feature-Based Models use deep learning technologies to autonomously learn complex, high-level features directly from vast amounts of data. This capability is essential for effectively addressing common challenges in tracking such as occlusions, rapid movements, and significant appearance changes of the target object. Numerous models have developed and implemented including GOTURN \cite{held2016learning}, DeepSORT \cite{yang2022video}, MDNet \cite{nam2016learning}, and DenseNet \cite{ma2020adaptive,lu2020dense}. These models are characterized by their ability to generalize across a variety of scenarios without requiring customized features, thus offering enhanced adaptability and robustness ideal for dynamic and real-time tracking applications.  They are also suitable in scenarios where their distinct capabilities, such as high-speed processing and adaptability to rapid appearance changes, which are essential for overcoming the limitations of traditional tracking approaches.

\subsection{Graph-Based Models}
\label{subsec:Graph-Based Models}
Graph-based models represent objects and their potential movements as nodes and edges in a graph to manage complex interactions and relationships between objects across video sequences. This structural representation allows the models to track individual identities and predict trajectories effectively, even in scenarios where objects frequently occlude each other. Key techniques in this area, such as K-Shortest Paths (KSP) Tracking \cite{berclaz2011multiple} and Dynamic Graph Matching \cite{zheng2011structured}, are efficient for maintaining a clear distinction between closely situated objects, enhancing tracking accuracy in densely populated settings. These models provide a structured means to disentangle the intricate movements of multiple objects, ensuring both high accuracy and reliability in tracking. However, the computational demands associated with updating and maintaining large-scale graphs pose challenges, particularly in real-time applications. Despite these challenges, graph-based models are well-suited to critical tracking environments, such as urban traffic systems and complex surveillance scenarios, where their systematic management of object trajectories proves indispensable.

\subsection{Data Association Models}
\label{subsec: Data Association Models}
Data association models facilitate the correlation of detection responses with object tracks—a process that grows increasingly complex as the number of objects and their interactions increases. They are categorized into Global Data Association Models \cite{schulter2017deep}, Online Data Association Models \cite{sahbani2016kalman}, and Probabilistic Data Association Models \cite{zulkifley2012robust}. Each model is tailored to specific challenges in MOT such as occlusions, dynamic interactions, and identity maintenance. This contributes to the precision and reliability of applications such as public transportation and automated surveillance.
For instance, Global models deal with comprehensive scene analysis, online models provide immediacy in decision-making, and probabilistic models offer robustness against detection uncertainties. However, these models also face limitations, such as computational complexity in global models, the need for rapid processing in online models, and the computational complexity of managing multiple hypotheses in probabilistic models. The selection of a particular model typically depends on the application's specific demands and constraints, with global models being ideal for batch-processed environments \cite{liu2005study}, online models suited for real-time applications \cite{wojke2017simple}, and probabilistic models best for settings with high uncertainty or crowded scenes \cite{li2014crowded}.

\section{Trajectory Prediction Models}
\label{sec:TP_MODELS}
Based on their theoretical foundations and implementation mechanisms, trajectory prediction models are divided into four classes: physics-based, classical machine learning, deep learning, and reinforcement learning methods. This classification provides essential insight into the strengths and limitations of each method, as well as facilitates the selection of the most appropriate techniques depending on specific application needs and environmental conditions.

\subsection{Physics-based Models}
\label{subsec:Physics-based Models}
In Early trajectory prediction tasks, physics-based models provide an intuitive understanding of motion dynamics, which makes them particularly suitable for preliminary predictions. These models can be categorized according to their underlying principles into three main types: Single Trajectory Prediction \cite{huang2022survey,lin2000vehicle, pepy2006reducing,kaempchen2009situation}, Kalman Filtering \cite{kaempchen2004imm}, and Monte Carlo \cite{okamoto2017driver}.
While these models offer simplicity and computational efficiency but they have difficulties in dealing with complex dynamics and environmental uncertainties. The limitations underscore the imperative for an integrative approach that combines physics-based models with machine learning to enhance prediction accuracy and adaptability. Such a fusion represents a promising direction for future research, aiming to significantly improve dynamic object tracking.

\subsection{Traditional Machine Learning Models}
\label{subsec:machine_learning}
Classical Machine learning Models shift trajectory prediction from physics-based to data-driven models, adeptly handling complex, nonlinear dynamics in object tracking. Techniques like Gaussian Process (GP) \cite{rasmussen2003gaussian}, Support Vector Machine (SVM) \cite{mandalia2005using}, Hidden Markov Model (HMM) \cite{qiao2014self}, and Dynamic Bayesian Network (DBN) \cite{murphy2002dynamic} capitalize on historical data for precise future trajectory predictions. This transition underscores machine learning's essential contribution to improving predictions in applications such as autonomous driving and pedestrian tracking, but they still have some significant limitations. uture research directions may include the development of hybrid models that synergize the strengths of machine learning with physics-based insights for more accurate, reliable, and context-aware prediction systems in dynamic object tracking.

\subsection{Deep Neural Network Models}
\label{subsubsec:deep_learning}
We have already discussed the impact of deep learning models on dynamic object tracking. These have also been applied to trajectory prediction. For instance graph-neural networks {GNNs}, which represent each object as a node within a graph, often leading to the representation of scenes as irregular graphs with variable-sized, unordered nodes. the connectivity among nodes via edges allows for capturing object inter-dependencies effectively. GNNs thus emerge as apt for tackling vehicle trajectory prediction issues rooted in interaction-related factors \cite{diehl2019graph}.

Despite their successes, challenges such as data dependency, computational demands, and the integration of multimodal data sources remain. Future research directions may include refining network architectures for greater efficiency, exploring unsupervised and semi-supervised learning paradigms to alleviate data labeling demands, and developing hybrid models that synergize deep learning with domain-specific knowledge. 

\subsection{Reinforcement Learning Models}
\label{subsec:hybrid_approaches}
Reinforcement Learning (RL) has emerged as a potent tool in trajectory prediction, offering a novel approach to understanding high-dimensional complex policies for Autonomous Vehicles (AVs) \cite{bharilya2023survey}. Utilizing the Markov Decision Process (MDP) framework, RL-based methods aim to maximize expected cumulative rewards through actions in given states, adapting to the dynamic nature of object tracking. This paradigm shift towards RL in trajectory prediction includes methodologies such as Inverse Reinforcement Learning (IRL) \cite{fernando2020deep}, Generative Adversarial Imitation Learning (GAIL) \cite{bhattacharyya2022modeling}, and Deep Inverse Reinforcement Learning (DIRL) \cite{xu2023solo}, each offering unique perspectives on learning from expert demonstrations to predict future trajectories.

\section{Datasets, Metrics, and Evaluation}
\label{sec:Performance Evaluation}
This section addresses Datasets, Metrics, and Evaluation, elucidating the difference between streamed and historical data to optimize dataset management and tailoring metric selections to each tracking type accordingly. It further delineates the metrics specific to DOT and TP, underpinning their relevance and applicability in assessing system performance. Additionally, this section evaluates tracking methods and trajectory prediction technologies highlighting the technological disparities between SOT and MOT and highlighting pivotal challenges and opportunities that could propel advancements in the field.

\subsection{Dateset}
\label{subsec:Dateset}
This research emphasizes the importance of using diverse datasets to ensure that models can perform well and be applied in different situations. Additionally, we suggest categorizing datasets into historical and streamed types, each with its own advantages. Historical data provides a strong foundation for learning from past events and patterns, which is crucial for training models to work well in similar future scenarios. On the other hand, streamed data offers real-time information, which is essential for models operating in dynamic environments, improving the accuracy and adaptability of tracking systems.

Historical and streamed data, both typically structured as time series, play distinct but complementary roles in modern data analysis and machine learning applications. Historical data consists of pre-collected, often pre-processed datasets that chronicle detailed metrics over substantial periods. This data type is structured and rich, providing deep insights into trends, patterns, and anomalies, which are invaluable for training algorithms to detect and interpret behaviours accurately. For instance, vast archives of surveillance footage serve as historical data, training systems to recognize various activities and behaviours over time. In contrast, streamed data is characterized by its real-time nature, continually captured, and immediately processed, typically from sensors or ongoing digital interactions. This volatile and less structured data form is vital for applications demanding quick analysis and responses, such as monitoring live video feeds from surveillance cameras. Streamed data enables systems to make swift, informed decisions and update predictions on the fly, allowing for effective adjustments to rapidly changing situations.

As illustrated in Figure 5, we systematically delineate the key differences between historical data and streamed data, focusing on their distinct approaches to data collection, preprocessing, and integration into machine learning (ML) models for dynamic object tracking and trajectory prediction.

\begin{figure}[h]
\centering
\includegraphics[width=0.9\textwidth]{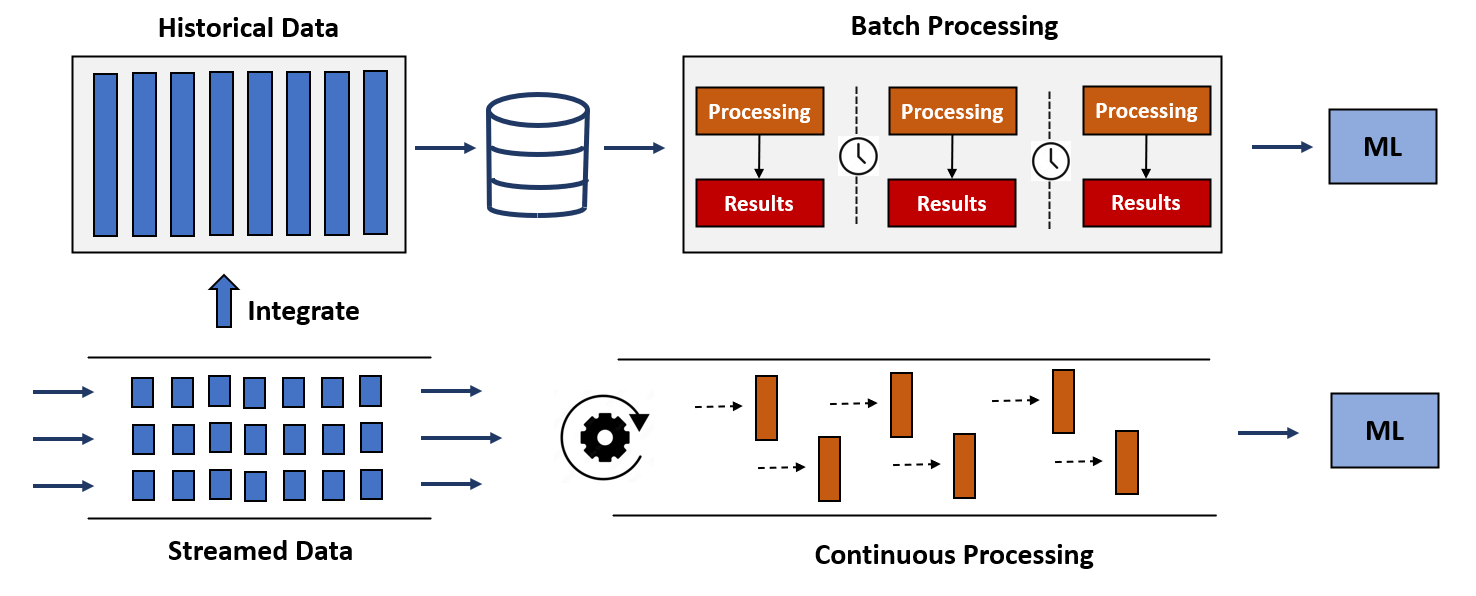}
\caption{Flowchart of Historical data and streamed data.}
\label{fig: flowchart of Historical and streamed data}
\end{figure}

\subsubsection{Data collection}
\label{subsec:Data collection}
The data collection in these systems is characterized by its continuous and real-time generation, often sourced from sensors, cameras, or online interactions. This data flows into the system in a live, unbroken stream, necessitating immediate handling to maintain its relevance for real-time applications. The collected data is less about storage and more about instant utility, focusing on current conditions and immediate insights. 

\subsubsection{Data preprocessing}
\label{subsec:Data preprocessing}
The preprocessing of data presents a significant challenge, exacerbated when dealing with streamed data as compared to historical data, especially in complex situations such as occlusion and rapid movement as shown in Table 1. Streamed data introduces additional complexities, such as live video feeds from cameras or real-time sensor data from mobile objects, which require immediate processing to ensure timely and effective object tracking.  The key preprocessing challenges and their corresponding technological solutions for both data types in DOT is outlined in Figure \ref{fig: DOT-TP Data Preprocessing Challenges}. Moreover, integrating data from multiple sensors also presents challenges, addressed through advanced data integration platforms that standardize and merge data inputs. 

\begin{figure}[h]
\centering
\includegraphics[width=0.9\textwidth]{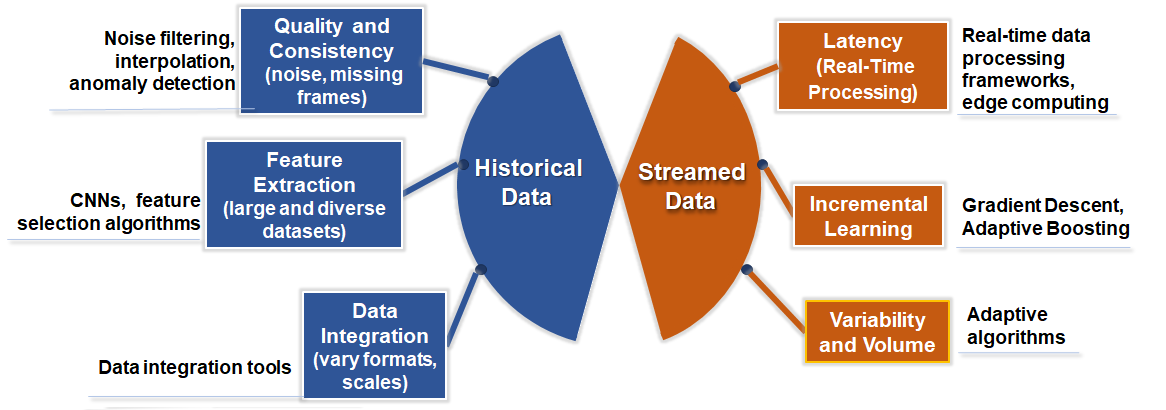}
\caption{Challenges and Technologies in Data Preprocessing for Dynamic Object Tracking: Historical vs. Streamed Data.}
\label{fig: DOT-TP Data Preprocessing  Challenges}
\end{figure}

\subsubsection{Dataset Selection and application}
\label{subsec:Dataset Selection and application}
Accurate dataset selection and classification play crucial roles in algorithm development and benchmarking. Table \ref{tab:Benchmarks} provides a comprehensive comparative analysis, categorizing leading datasets like KITTI, VOT2018, and Waymo by key attributes—data type, object types, volume, scenarios, resolution, and annotation quality. This structured classification aids in pinpointing datasets that align with specific research requirements, crucial for training advanced models and enhancing system sophistication. These datasets encompass a broad spectrum of scenarios and objects, ranging from urban vehicles and pedestrians to UAVs in natural settings, offering diverse challenges that broaden the applicability and relevance of research outcomes in real-world environments.

Table \ref{tab:Benchmarks} distinguishes between Historical Data and Streamed Data, with each category serving distinct analytical and application-oriented purposes within DOT frameworks. Historical Data includes rich, high-resolution datasets such as KITTI and the more recent MOTChallenge2023, designed for offline analysis. These datasets facilitate deep pattern analysis and comprehensive model training across various scenarios from urban to synthetic environments, demonstrating an evolution in dataset complexity and utility. On the contrary, Streamed Data caters to real-time applications like traffic monitoring and autonomous navigation, featuring datasets such as Drone Video Dataset and Public Transport Systems Dataset. These datasets are characterized by ongoing data collection and variable volumes, tailored for environments that demand instantaneous data processing and dynamic adaptability. This methodical organization of datasets not only meets diverse research demands but also underscores the significance of dataset classification in enhancing the accuracy and adaptability of DOT systems. This strategic approach to dataset utilization underscores a clear trajectory of development and expansion in dataset application scope, pivotal for advancing DOT capabilities in various operational contexts.

\begin{sidewaystable}
\centering
\caption{Benchmarks of Dynamic Object Tracking and Trajectory Perdition}
\label{tab:Benchmarks}
\renewcommand{\arraystretch}{1}
\begin{tabularx}{\textwidth}{@{} >{\raggedright\arraybackslash}p{1.3cm} >{\raggedright\arraybackslash}p{1.9cm}>
{\raggedright\arraybackslash}p{2.1cm}>
{\raggedright\arraybackslash}p{2.2cm}>
{\raggedright\arraybackslash}p{2.4cm}>
{\raggedright\arraybackslash}p{1.5cm}>
{\raggedright\arraybackslash}p{1.6cm}>
{\raggedright\arraybackslash}p{1.2cm}>
{\raggedright\arraybackslash}p{1.3cm} @{}}
\toprule
{\bf Data Type} & {\bf Dataset} & {\bf Object Types} & {\bf Volume} & {\bf Scenarios} & {\bf Resolution} & {\bf Annotation Quality} & {\bf Year} & {\bf Reference} \\

\midrule

\multirow{13}{=}{Historical Data (Offline analysis)}  
& KITTI & Vehicles, pedestrians & Thousands of frames & Urban driving scenes & High & High & 2012 &  \cite{geiger2013vision}\\

 & OTB-2015 & Single objects & 100 videos & Limited scenarios & High & High & 2015 &  \cite{danelljan2015learning}\\
 
 & VOT2016 & Single objects & 60 sequences & Controlled scenarios & High & High & 2016 & \cite{roffo2016visual} \\
 
 & Charades & Human activities & 9,848 videos & Home environments & High & High & 2016 & \cite{alamri2019audio} \\
 & UAV123 & UAVs, various small objects & 123 videos & Diverse natural environments & High & High & 2016 & \cite{mueller2016benchmark}\\
 & Stanford Drone & Pedestrians, cyclists, cars & Thousands of trajectories & Campus scenes with multiple a & High & Medium & 2016 & \cite{andle2022stanford} \\
 & MOT17 & Pedestrians & 14 sequences, thousands of frames & Various urban scenarios & High & High & 2017 & \cite{wang2019exploit} \\
 & VOT2018 & Single objects & 60 sequences & More challenging & High & High & 2018 &\cite{kristan2018sixth}  \\
 & GOT-10k & Over 563 object types & 10k+ videos, 1.5 million frames & Diverse environments & High & High & 2019 & \cite{huang2019got} \\
 & LaSOT & Single objects & 1,400 videos, hundreds of frames each & Diverse environments & High & High & 2019 &  \cite{fan2021lasot}\\
 & nuScenes & Vehicles, pedestrians & 1,000 scenes, 1.4M images & Various urban scenarios & High & High & 2019 & \cite{caesar2020nuscenes} \\
 & MOT20 & Pedestrians & 8 sequences & Dense crowds & High & High & 2020 & \cite{dendorfer2020mot20} \\
 & TAO & Multiple object types & 2,907 videos, 833 object classes & Diverse environments & High & High & 2020 & \cite{dave2020tao} \\
 & MOTSynth & Pedestrians & Millions of frames & Varied synthetic scenarios & High & High & 2021 & \cite{fabbri2021motsynth} \\
 & MOT Challenge & Pedestrians, vehicles & Multiple sequences & Varied urban scenarios & High & High & 2023& \cite{dendorfer2021motchallenge} \\
 & Waymo open dataset& Vehicles, pedestrians & Thousands of segments & Diverse urban and suburban scenes & High & Very High & 2024 & \cite{schwall2020waymo}\\
\midrule
\multirow{3}{=}{Streamed Data (Real-time Analysis)} 

 & Drone Video Dataset & Landscapes, vehicles, people & Potentially very large & Varied environments & High & Medium & Ongoing& \cite{perera2019drone} \\
 
 & Mobile Device Location Streams & Mobile devices locations & Extremely large (continuous stream) & Global coverage & Medium & Medium & Ongoing & \cite{karagkioules2018public}\\
 
 & Public Transport Systems Dataset & Buses, trains, stations & Large (continuous updates) & Focused on transit routes & Medium & Medium & Ongoing & \cite{kujala2018collection}\\

\bottomrule

\end{tabularx}
\end{sidewaystable}

\subsection{Metrics}
\label{subsec:Metrics}
The application of distinct metrics tailored for single object tracking, multiple object tracking, and trajectory prediction is indispensable. These metrics serve to comprehensively evaluate the performance of algorithms, each addressing specific aspects of accuracy, fairness, robustness, and calibration. Table \ref{tab:metrics} provides a comprehensive summary of Metrics for Evaluating Accuracy, Fairness, Robustness, and Calibration in Single Object Tracking (SOT), Multiple Object Tracking (MOT), and Trajectory Prediction (TP).

\begin{sidewaystable}
\centering
\caption{Comprehensive Overview of Metrics for Evaluating Accuracy, Fairness, Robustness, and Calibration in Single Object Tracking (SOT), Multiple Object Tracking (MOT), and Trajectory Prediction (TP)}

\label{tab:metrics}
\renewcommand{\arraystretch}{0.85}
\begin{tabularx}{1.03\textwidth}{@{} >{\raggedright\arraybackslash}p{1.35cm} |>{\raggedright\arraybackslash}p{6.0cm} |>
{\raggedright\arraybackslash}p{5.4cm}|>
{\raggedright\arraybackslash}p{5.95cm} @{}}
\toprule
{\bf Metrics} & {\bf Single Object Tracking (SOT)} & {\bf Multiple Object Tracking (MOT)} & {\bf Trajectory Prediction (TP)}  \\

\midrule

\multirow{3}{=}{Accuracy}  
 & Intersection over Union (IoU):Overlap between predicted and ground truth bounding boxes. \( \text{IoU} = \frac{\text{Area of Overlap}}{\text{Area of Union}} \),  
&  Multiple Object Tracking Accuracy (MOTA): Combines false positives, false negatives, and identity switches. \( \text{MOTA} = 1 - \frac{\sum_{t}(FN_t + FP_t + IDSW_t)}{\sum_{t}GT_t} \) 
&  Average Displacement Error (ADE): Average Euclidean distance between predicted and ground truth positions.  \( \text{ADE} = \frac{1}{T} \sum_{t=1}^T \sqrt{(x_t^{pred} - x_t^{gt})^2 + (y_t^{pred} - y_t^{gt})^2}\)  \\

& Precision: Average distance between centers of predicted and ground truth bounding boxes. \( \text{Precision} = \frac{1}{N} \sum_{i=1}^N \sqrt{(x_i^{pred} - x_i^{gt})^2 + (y_i^{pred} - y_i^{gt})^2} \),   
& Multiple Object Tracking Precision (MOTP): Alignment of predicted and ground truth bounding boxes for matched objects. \( \text{MOTP} = \frac{\sum_{i,t} d_{i,t}}{\sum_{t} c_t} \)
&  Final Displacement Error (FDE), Euclidean distance between predicted and ground truth positions at final time step.  \(\text{FDE} = \sqrt{(x_T^{pred} - x_T^{gt})^2 + (y_T^{pred} - y_T^{gt})^2}\) \\
& Success Rate: Proportion of frames where IoU is above a certain threshold.  \( \text{Success Rate} = \frac{\sum_{i=1}^N \mathbf{1}(\text{IoU}_i > \text{threshold})}{N} \),  
& ID Switches (IDSW): Number of times an object’s identity is incorrectly assigned.
& \\
       
\midrule

\multirow{3}{=}{Fairness}  
&Bias Analysis: Difference in tracking performance across object categories. \( \text{Bias} = \frac{1}{N} \sum_{i=1}^N \left(P_{i, \text{group}} - P_{i, \text{overall}}\right) \) 

& Demographic Parity: Similar tracking accuracy across demographic groups. \( \text{Demographic Parity} = \frac{P(\text{Success} | \text{Group A})}{P(\text{Success} | \text{Group B})} \)
& Prediction Bias: Deviation in prediction accuracy across demographic groups. \( \text{Prediction Bias} = \frac{1}{N} \sum_{i=1}^N \left(\text{Error}_{i, \text{group}} - \text{Error}_{i, \text{overall}}\right) \) 
\\

& Error Distribution: Distribution of tracking errors across conditions. 

& Consistency Across Scenes: Performance consistency across different scenes
& Calibration Fairness: Prediction reliability across different groups.
\\

& Equality of Opportunity: Equal opportunity for success across object groups. \( \text{Equality of Opportunity} = \frac{\text{TPR}_{\text{group}}}{\text{TPR}_{\text{overall}}} \) 
& Disparate Impact: Impact of tracking system on different groups. \( \text{Disparate Impact} = \frac{\text{Error Rate}_{\text{group}}}{\text{Error Rate}_{\text{overall}}} \) 
& Equalized Odds: Similar error rates across different groups. \( \text{Equalized Odds} = \frac{\text{FPR}_{\text{group}}}{\text{FPR}_{\text{overall}}} \) 
\\

\midrule

\multirow{3}{=}{Robustness}  
& Occlusion Handling:Performance under occlusion.\( \text{Occlusion Robustness} = \frac{\text{IoU}_{\text{occluded}}}{\text{IoU}_{\text{unoccluded}}} \) 

& Density Handling: Performance in scenes with different object densities.\( \text{Density Robustness} = \frac{\text{MOTA}_{\text{high density}} + \text{MOTA}_{\text{low density}}}{2} \)  &  Trajectory Change Handling: Prediction accuracy under sudden trajectory changes.\( \text{Trajectory Change Robustness} = \frac{\text{ADE}_{\text{changed}}}{\text{ADE}_{\text{unchanged}}} \)\\

&  Illumination Variation: Performance under varying lighting conditions.\( \text{Illumination Robustness} = \frac{\text{IoU}_{\text{low light}} + \text{IoU}_{\text{bright light}}}{2} \)

& Motion Robustness:Performance with objects moving at different speeds.\( \text{Motion Robustness} = \frac{\text{MOTA}_{\text{fast}} + \text{MOTA}_{\text{slow}}}{2} \) 

&  Environmental Robustness: Prediction performance under different environmental conditions. \( \text{Environmental Robustness} = \frac{\text{ADE}_{\text{condition 1}} + \text{ADE}_{\text{condition 2}}}{2} \)\\

& Deformation Handling: Performance under object shape changes. \( \text{Deformation Robustness} = \frac{\text{IoU}_{\text{deformed}}}{\text{IoU}_{\text{original}}} \)   

&  Occlusion Handling:Performance under frequent occlusions. \( \text{Occlusion Robustness} = \frac{\text{MOTA}_{\text{occluded}}}{\text{MOTA}_{\text{unoccluded}}} \) 
& Long-term Prediction: Accuracy of long-term predictions. \( \text{Long-term Robustness} = \frac{\text{FDE}_{\text{long-term}}}{\text{FDE}_{\text{short-term}}} \)  \\
 
 \midrule

 \multirow{2}{=}{Calibration}  
&  Expected Calibration Error (ECE): Difference between predicted confidence and actual accuracy. \( \text{ECE} = \sum_{m=1}^M \frac{|B_m|}{n} \left| \text{acc}(B_m) - \text{conf}(B_m) \right| \)  & Expected Calibration Error (ECE): Difference between predicted confidence and actual accuracy for multiple objects. \( \text{ECE} = \sum_{m=1}^M \frac{|B_m|}{n} \left| \text{acc}(B_m) - \text{conf}(B_m) \right| \) 

&  Expected Calibration Error (ECE): Difference between predicted confidence and actual accuracy for predicted trajectories. \( \text{ECE} = \sum_{m=1}^M \frac{|B_m|}{n} \left| \text{acc}(B_m) - \text{conf}(B_m) \right| \) \\

& Brier Score: Mean squared difference between predicted probabilities and actual outcomes. \( \text{Brier Score} = \frac{1}{n} \sum_{i=1}^n (p_i - o_i)^2 \) 
& Log-Loss:Performance measure where the prediction input is a probability value between 0 and 1. \( \text{Log-Loss} = -\frac{1}{n} \sum_{i=1}^n [o_i \log(p_i) + (1 - o_i) \log(1 - p_i)] \) 

&   Negative Log-Likelihood (NLL): Likelihood of the predicted trajectory under the true distribution. \( \text{NLL} = -\sum_{i=1}^n \log P(o_i|p_i) \)  \\

\bottomrule

\end{tabularx}
\end{sidewaystable}

\subsubsection{Accuracy}
Accuracy refers to the closeness of the predicted outputs to the ground truth data. It is fundamental in assessing the effectiveness of tracking and prediction models. High accuracy ensures that the system can reliably follow or predict the path of an object across frames or time steps. Common metrics for evaluating accuracy include Precision, Recall, Intersection over Union (IoU)\cite{rezatofighi2019generalized}, and Mean Squared Error (MSE) \cite{koksoy2006multiresponse}. These are applied in single object tracking to measure individual tracking precision, in multiple object tracking to handle overlapping and identity switches, and in trajectory prediction to assess future path accuracy.

Accuracy in Dynamic Object Tracking (SOT or MOT) and Trajectory Prediction (TP) is evaluated using specific metrics tailored to the tracking complexity and prediction requirements. For instance IoU, which calculates the overlap between between the predicted bounding box and the ground truth, is heavily used in DOT \cite{wang2020towards,luo2018fast,zhao2021adaptive,jiao2021deep}. Other metrics e.g., (MOTA) combine false positives, false negatives and identity switches to provide an overall score with higher values denoting better performance.

In TP, accuracy is assessed using Average Displacement Error (ADE) and Final Displacement Error (FDE). ADE measures the average Euclidean distance between predicted and actual positions over all time steps, with lower values indicating higher prediction accuracy. FDE evaluates the Euclidean distance between the predicted and actual positions at the final time step, assessing long-term prediction accuracy. These metrics collectively ensure comprehensive evaluation of tracking and prediction systems, enhancing their reliability and performance in applications like surveillance, autonomous driving, crowd analysis, and robotic path planning.

\subsubsection{Fairness}
Fairness in object tracking and trajectory prediction ensures that the performance of algorithms is consistent across different subgroups or scenarios. This metric is crucial to avoid biases based on object size, speed, or environmental conditions, which can skew the effectiveness of the application in diverse real-world settings. Metrics like Equality of Opportunity \cite{roemer2015equality} and Demographic Parity\cite{jiang2022generalized} are employed primarily in multiple object tracking, where diverse object categories and conditions are present, ensuring uniform performance across various groupings. 

In TP, fairness is gauged through Prediction Bias, Calibration Fairness, and Equalized Odds. Prediction Bias measures deviations in prediction accuracy across demographic groups. Calibration Fairness assesses the reliability of predicted trajectories across groups. Equalized Odds measure the similarity in error rates across groups.

\subsubsection{Robustness}
Robustness measures the ability of algorithms to maintain high performance under different operating conditions. This includes challenges such as varying lighting, occlusions, and camera angles. For single object tracking, robustness metrics focus on tracking stability despite object appearance changes or occlusions. For multiple object tracking, they evaluate the system's ability to handle scene complexity and dynamic interactions. Adversarial Robustness, Performance under Occlusion, and Weather and Lighting Variance are key metrics. In trajectory prediction, robustness ensures reliability even with unpredictable movements or environmental changes.

\subsubsection{Calibration}
Calibration evaluates how well the probability outputs of a model reflect the actual likelihood of predictions. It is particularly crucial in trajectory prediction, where confidence in the predicted paths influences decision-making in applications like autonomous driving. Properly calibrated models offer reliable predictions that stakeholders can trust. Metrics such as Expected Calibration Error (ECE)\cite{cao2024cctr}, Calibration curves \cite{moosavi2018linearity}, and the Brier Score \cite{feng2023macformer} are used to assess this aspect, ensuring the predictions' reliability and accuracy over time.

These metrics are systematically applied across the specific tasks of single object tracking, multiple object tracking, and trajectory prediction to refine algorithms and enhance their performance, addressing the nuanced demands of each application area. The critical evaluation using these metrics not only aids in benchmarking current methods but also in driving advancements in dynamic object tracking technologies.

\subsection{Evaluation}
\label{subsec:Evaluation}
This section evaluates related studies of trajectory prediction in five scenarios, each characterized by particular performance metrics: accuracy, fairness, robustness, and calibration, as summarized in table \ref{tab:single_object_tracking}. The evaluation delves into scenarios ranging from Single Object Tracking (S1) to complex Multi-Object Tracking with Multiple SOT Algorithms (S5), highlighting the nuanced challenges and interactions inherent to each based on these performance metrics. We analyze each model's performance under varying complexity and predictability conditions and their strengths, pinpoint critical research gaps, and provide strategic insights for future applications based on these findings.

\begin{sidewaystable}
\centering
\caption{Evaluation of Object Tracking for Trajectory Prediction in Five Scenarios.  S1: Single Object Tracking. S2: MOT (Known \& Identified Object Tracking). S3: MOT (Class-Specific Object Tracking). S4: MOT (Unknown Objects \& Classes). S5: MOT (Multi-Object Tracking with Multiple SOT Algorithms).}
\label{tab:single_object_tracking}
\renewcommand{\arraystretch}{0.95}
\begin{tabularx}{1.01\textwidth}{@{} >{\centering\arraybackslash}p{1.05cm}>
{\raggedright\arraybackslash}p{1.05cm}>
{\raggedright\arraybackslash}p{2.8cm}>
{\raggedright\arraybackslash}p{2.95cm}>
{\raggedright\arraybackslash}p{2.85cm}>
{\centering\arraybackslash}p{1.2cm}>
{\centering\arraybackslash}p{1.2cm}>
{\centering\arraybackslash}p{1.4cm}>
{\centering\arraybackslash}p{1.4cm}@{}}

\toprule

{\bf Scenarios} & {\bf Studies} & {\bf DOT Model} & {\bf TP Model} & {\bf Dataset}  & {\bf Accuracy} & {\bf Fairness} & {\bf Robustness} & {\bf Calibration}\\ 

\midrule

S1 & \cite{weser2006multimodal} & Sensor-based multimodal & Self Organizing Maps for motion pattern learning & SICK lasers &  $\checkmark$ & $\times$ & $\checkmark$ & $\times$  \\ 

S1 & \cite{wang2017trajectory} & CNN & RNN with LSTM & Online Vehicle dataset &  $\checkmark$ & $\times$ & $\times$ & $\times$    \\

S1 & \cite{wu2020visual} & Collaborative Correlation Filter & Cross-camera TP Network & GMTD and CAMPUS datasets &   $\checkmark$ & $\times$ & $\checkmark$ & $\times$  \\ 

S1 & \cite{wang2021dynamic} & Dynamic Attention Guided Tracking & Multi-Trajectory Selection Network & OTB2015, GOT-10k, UAV123, etc. &   $\checkmark$ & $\times$ & $\times$ & $\times$    \\

S1 & \cite{xu2022non} & Siamese Network & LSTM Trajectory Prediction via GAN  & OTB2015, VOT2016\&2018 &   $\checkmark$ & $\times$ & $\checkmark$ & $\times$  \\ 

\hline

S2 & \cite{leibe2007coupled} & Hypothesis selection framework & Model selection framework, non-Markovian & Custom surveillance videos &   $\checkmark$ & $\times$ & $\times$ & $\times$  \\ 

S2 & \cite{fernando2018tracking} & GANs for person detection & Trajectory prediction with LSTMs & PETS2009, AVG-TownCentre &  $\checkmark$ & $\times$ & $\times$ & $\times$  \\

S2 & \cite{li2019multi} & Multi-target tracking with LSTM & LSTM with attention & MOT16, MOT17 &   $\checkmark$ & $\times$ & $\checkmark$ & $\times$   \\

S2 & \cite{girbau2021multiple} & Multiple Object Tracking with TrajE & Mixture Density Networks  & MOT-Challenge, UA-DETRAC &  $\checkmark$ & $\times$ & $\times$ & $\times$  \\ 

S2 & \cite{dendorfer2022quo} & Monocular multi-object tracking (MOT) & Trajectory forecasting in BEV space &  MOTSynth, MOT17\&20 &  $\checkmark$ & $\times$ & $\checkmark$ & $\times$   \\

S2 & \cite{jeon2023leveraging}& Deep neural networks  & Machine learning models & MOT-Challenge, TownCenter &   $\checkmark$ & $\times$ & $\times$  & $\times$  \\

\hline

S3 &  \cite{fuchs2019end} & MOHART & Self-attention & MOT-Challenge, UA-DETRAC &  $\checkmark$ & $\times$ & $\times$ & $\times$  \\ 

S3 & \cite{wu20213d} & Prediction Confidence-Guided Data Association & Constant Acceleration motion model & KITTI &  $\checkmark$ & $\times$ & $\checkmark$ & $\times$   \\

S3 & \cite{weng2022whose} & Affinity-based detections & Transformer with affinity attention & Autonomous driving datasets &   $\checkmark$ & $\times$ & $\checkmark$ & $\times$  \\ 

S3  &  \cite{zhang2023multi} & Contextual features, graph model & Kalman filters, GNN & KITTI, nuScenes &   $\checkmark$ & $\times$ & $\times$  & $\times$   \\

\hline

S4 & \cite{wan2021multiple} & Temporal Priors Embedding  & Trajectory Map Regression & MOT16\&17\&20 &  $\checkmark$ & $\times$ & $\checkmark$ & $\times$   \\

S4 & \cite{kesa2021joint} & Joint Learning Architecture & Embedded forecasting network & MOT-Challenge, MOT16\&17\&20 &   $\checkmark$ & $\times$ & $\times$ & $\times$  \\ 

S4 &  \cite{chen2023trajectoryformer} & Transformer-based 3D MOT framework  & Predictive Trajectory Hypotheses  & Waymo3DMOT &  $\checkmark$ & $\times$ & $\times$ & $\times$  \\ 

\hline

S5 & \cite{he2017sot} & Deep learning based appearance model  & Appearance model & MOT16 &  $\checkmark$ & $\times$ & $\checkmark$ & $\times$   \\

S5 & \cite{liang2021generic} & Integration of SOT, Tracklet, and Re-identification & Spatial Attention and Hierarchical Clustering & MOT16\&17 &   $\checkmark$ & $\times$ & $\checkmark$  & $\times$  \\ 

S5 &  \cite{zheng2021improving} & Integrated SOT branch, extending CenterNet & Online discriminative training  & MOT17\&20 &  $\checkmark$ & $\times$ & $\checkmark$  & $\times$  \\ 

S5 &  \cite{zhang2023bidirectional} & Bidirectional MOT with SOT Integration & Trajectory Criteria  & VISO, SkySat-1 &  $\checkmark$ & $\times$ & $\checkmark$ & $\times$  \\

\bottomrule

\end{tabularx}
\end{sidewaystable}

\subsubsection{Scenario 1: Single Object Tracking (SOT)}
\label{sec:Scenario 1}
SOT involves the continuous observation and tracking of a singular identified object within a video stream, focusing primarily on predicting its trajectory based on past motion patterns and visual cues. 

These approaches utilize various datasets like SICK lasers, OTB2015, and VOT2018, and metrics including precision, recall, and F1-Score to validate their effectiveness. The strengths of these systems often include enhanced robustness and accuracy in challenging conditions such as occlusions, while weaknesses typically involve high computational loads and the dependency on initial tracking accuracy, which can degrade performance. 

Particularly, studies like Lituan et al. \cite{wang2017trajectory} and Wang et al. \cite{wang2021dynamic} highlight the trade-offs between real-time performance capabilities and computational demands. Despite advancements, key gaps remain in the areas of fairness, robustness against diverse tracking scenarios, and calibration of the tracking systems to improve their generalizability and reliability.

\subsubsection{Scenario 2: MOT (Known \& Identified Object Tracking)}
\label{sec:Scenario 2}
In the realm of MOT within the Known \& Identified Object Tracking scenario, researchers face the complex task of tracking multiple distinct objects simultaneously, each marked by a unique identifier. This scenario has many applications like team sports analysis, where understanding each player's specific movement patterns can profoundly influence strategic decisions and performance evaluations. The primary challenge lies in effectively managing potential interactions and overlaps among these objects, addressing issues such as collisions, occlusions, and maintaining accurate data association and identity preservation over time. Studies such as those by Leibe et al. \cite{leibe2007coupled} implemented a hypothesis selection framework to improve trajectory recovery, although their effectiveness was somewhat limited by the capabilities of the underlying object detection systems.

Further developments in MOT technology illustrate continuous efforts to refine tracking accuracy and adaptability to complex environments. For instance, Girbau et al. \cite{girbau2021multiple} introduced a lightweight model using Multiple Object Tracking with TrajE and Mixture Density Networks, which improved occlusion handling but faced challenges such as potential overfitting in dynamic scenarios. Additionally, recent innovations by Dendorfer et al. \cite{dendorfer2022quo} presented a monocular MOT system that employs trajectory forecasting in Bird’s Eye View (BEV) space, successfully reducing identity switches and enhancing robustness against prolonged occlusions but at the cost of increased model complexity and sensitivity to depth estimates. Jeon et al. \cite{jeon2023leveraging} pushed the boundaries further by leveraging deep neural networks to significantly boost tracking accuracy and reduce identity switches, although their methods necessitated high computational loads and were constrained by specific camera setups. Despite these advancements, significant gaps remain in terms of real-time performance, system calibration, and operational flexibility, highlighting the ongoing need for research and development. Addressing these persistent challenges, such as the high computational demands, extensive training requirements, and sensitivity to initial conditions, is crucial for enhancing the practical deployment and reliability of MOT systems in dynamically complex environments.

\subsubsection{Scenario 3: MOT (Class-Specific Object Tracking)}
\label{sec:Scenario 3}
Class-Specific Object Tracking (MOT) addresses the challenge of monitoring multiple objects not categorized as individuals but as members of specific classes. This tracking paradigm is essential for applications such as urban traffic control, where the ability to predict and manage the flow of similar vehicle types can significantly optimize traffic signal coordination and congestion management. The primary challenges in this approach include handling the dynamics of interactions within and between classes, managing overlaps, and resolving intra-class occlusions. These elements are crucial for accurately modeling and predicting the generalized movement patterns typical of each class, which are fundamental for effective system implementation in real-world dynamic environments.

Fuchs et al. \cite{fuchs2019end} applied MOHART with self-attention mechanisms to the MOT-Challenge and UA-DETRAC datasets, achieving improved intersection over union (IoU) metrics but at the cost of high computational demands. Wu et al. \cite{wu20213d} enhanced tracking accuracy and robustness using a Prediction Confidence-Guided Data Association system with a Constant Acceleration motion model on the KITTI dataset, though it faced difficulties with sparse point clouds. Weng et al.\cite{weng2022whose} utilized affinity-based detections with a Transformer model focused on affinity attention, significantly reducing prediction errors in autonomous driving datasets but dependent on the quality of affinity matrices. 

Additionally, Zhang et al. \cite{zhang2023multi} integrated contextual features and graph models with Kalman filters and Graph Neural Networks (GNNs) on KITTI and nuScenes, achieving high MOTA scores and reduced identity switches, albeit struggling in densely populated scenarios. These developments indicate a positive trajectory in the field of class-specific MOT, yet underline existing gaps such as the need for real-time performance, system calibration, and operational flexibility. Addressing these challenges is crucial, as the success of MOT systems in real-world applications depends heavily on their ability to efficiently and accurately process complex, dynamic scenarios.

\subsubsection{Scenario 4: MOT (Unknown Objects \& Classes)}
\label{sec:Scenario 4}
MOT for Unknown Objects \& Classes, tackles the challenge of detecting and monitoring objects without predefined knowledge of their number or classes. This scenario can be found in autonomous driving, where systems must dynamically identify and track a variety of objects—such as pedestrians, vehicles, and animals—that enter and exit the sensor's range unpredictably. The core challenge lies in the systems' ability to manage the uncertainty of detection and to adaptively learn and predict object movements as new data emerges, necessitating algorithms capable of robustly handling interactions based on spatial relationships. 

Recent, Wan et al. \cite{wan2021multiple} implemented Temporal Priors Embedding with Trajectory Map Regression, achieving high accuracy and effective occlusion management on datasets like MOT16, MOT17, and MOT20, though heavily reliant on precise initial detection. Kesa et al. \cite{kesa2021joint} developed a Joint Learning Architecture that significantly reduced identity switches but faced challenges with high false positives in complex scenarios. Meanwhile, Chen et al. \cite{chen2023trajectoryformer} introduced a transformer-based 3D MOT framework that incorporated Predictive Trajectory Hypotheses, setting new benchmarks in performance on the Waymo3DMOT dataset but with substantial computational demands. These studies collectively push forward the capabilities in tracking unknown objects and classes; however, they underscore ongoing gaps such as high dependency on initial detection accuracy, challenges in handling occlusions, and the substantial computational resources required, emphasizing the need for further advancements in system robustness, calibration, and real-time processing efficiency.

\subsubsection{Scenario 5: MOT (Multi-Object Tracking with Multiple SOT Algorithms)}
\label{sec:Scenario 5}
It involves utilizing a variety of SOT algorithms to independently track multiple objects, each potentially employing a different tracking model. This approach is crucial in applications such as multi-camera surveillance systems in large public areas, where accurately predicting the paths of numerous individuals simultaneously is essential for effective security monitoring and incident response. The primary challenge lies in the coordination among disparate SOT algorithms to ensure seamless integration of tracking outputs and the efficient allocation of computational resources. 

He et al. \cite{he2017sot} utilized a deep learning-based appearance model on the MOT16 dataset, enhancing tracking accuracy but facing calibration limitations. Liang et al.  \cite{liang2021generic} merged SOT, tracklet connectivity, and re-identification techniques, employing spatial attention and hierarchical clustering to achieve robust tracking on the MOT16 and MOT17 datasets, although challenges in maintaining fairness among tracked objects were noted. Zheng et al.  \cite{zheng2021improving} integrated an SOT branch into CenterNet and applied online discriminative training on the MOT17 and MOT20 datasets, substantially improving robustness. Zhang et al.\cite{zhang2023bidirectional} implemented a bidirectional MOT system integrating SOT, which refined trajectory predictions on datasets like VISO and SkySat-1 but raised concerns over computational efficiency. While these developments mark significant progress, substantial gaps persist in optimizing accuracy, fairness, robustness, and system calibration, underscoring the ongoing need for innovative solutions to enhance the integration and operational efficiency of using multiple SOT algorithms in complex tracking environments.

\subsection{Discussion}
\label{sec:Discussion}
Combining DOT and TP represents a significant evolution in the capabilities of real-time tracking systems, particularly when addressing the challenges inherent to SOT and MOT. This systematic review has critically evaluated the synergistic integration of DOT and TP, revealing substantial advancements while also identifying persistent gaps and future research opportunities.

The integration of DOT and TP showcases a compelling approach to enhancing the precision and reliability of tracking systems. By combining the real-time tracking capabilities of DOT with the predictive insights of TP, systems can anticipate and adapt to object movements more effectively. This integration is especially critical in environments where swift and accurate tracking is paramount, such as in autonomous vehicle navigation and urban surveillance.

However, our review also highlights several key challenges that persist in the field. The issues, such as occlusion, the variability of object speeds and sizes, and environmental factors like lighting and weather conditions, continue to pose significant obstacles to the accuracy of DOT systems. These challenges are compounded in MOT scenarios where multiple objects interact dynamically, increasing the complexity of tracking and prediction tasks.

This study identifies a notable gap in the current literature concerning the long-term efficacy of integrated DOT and TP systems. While short-term improvements are well-documented, the sustainability of these enhancements over longer periods and under varying conditions remains under-explored. 

In addition, our examination of the current methodologies, models, datasets, and metrics highlights the need for more comprehensive and standardized evaluation methods. Currently, the field suffers from a fragmentation of approaches, making comparative analysis and the synthesis of findings challenging. Establishing more uniform metrics and evaluation protocols could facilitate more meaningful comparisons between studies and accelerate the advancement of the field.

Finally, the review underscores the importance of context-aware and privacy-preserving technologies in the advancement of DOT systems. As tracking technologies become more pervasive, ensuring they operate within ethical boundaries and respect user privacy is paramount. Future research should thus not only focus on enhancing technical capabilities but also on ensuring these technologies are implemented responsibly and ethically.

\section{Future Directions}
\label{subsec:Future Directions}
The convergence of dynamic object tracking and trajectory prediction with emerging technologies opens up new frontiers for research. This survey identifies several pivotal directions that promise to catalyze significant advancements in the field, informed by the comprehensive analysis of existing methodologies, applications, and challenges.

\paragraph*{Integration of Multimodal Data Sources} Enhancing dynamic object tracking across varied scenarios necessitates the exploration of multimodal data integration, spanning visual, auditory, and other contextual information. Future research should pioneer advanced algorithms for the seamless fusion of these diverse data streams, aiming to bolster the robustness and adaptability of tracking systems in complex environments.

\paragraph*{Semantic Information Fusion} The development of sophisticated fusion techniques that amalgamate semantic information from multiple modalities stands as a critical research avenue. Such techniques would not only refine tracking decisions but also ensure dynamic adaptability to fluctuating environmental contexts, significantly elevating tracking accuracy and relevance.

\paragraph*{Context-aware Dynamic Object Tracking} The ambition to design a novel multimodal semantic fusion framework underscores the need for context-aware dynamic object tracking solutions. Such frameworks would leverage the integrated information from various modalities, offering unprecedented accuracy and situational responsiveness in object-tracking applications.

\paragraph*{Advanced Integration} Integrating advanced tracking solutions like the THOR tracker with fuzzy logic and a multi-modal semantic dynamic model. This combination aims to leverage THOR's precise tracking capabilities with fuzzy logic's adaptability and decision-making efficiency, enhancing performance in complex and uncertain environments. Additionally, incorporating a multi-modal semantic model could improve systems' understanding and anticipation of object movements by analyzing rich contextual information. 

\paragraph*{Privacy-preserving Data Sharing} As the integration of data across modalities intensifies, designing mechanisms that safeguard privacy while enabling effective data sharing becomes paramount. Future efforts should focus on the creation of innovative privacy-preserving frameworks that facilitate cross-modal data utilization without compromising sensitive information.

\paragraph*{Comprehensive Performance Evaluation} Rigorous evaluation methodologies are essential to validate the efficacy of proposed approaches. Future research should encompass extensive experimentation and real-world scenario testing, utilizing diverse datasets to comprehensively assess performance trade-offs between tracking accuracy, context awareness, and privacy preservation.

\paragraph*{Efficient and Real-time Processing}  the optimization of tracking algorithms for deployment on edge computing platforms represents a future direction with profound implications. Emphasizing real-time processing capabilities and reduced computational demands will be crucial for applications demanding instantaneous decision-making.

\paragraph*{Ethical and Societal Implications} Finally, the pervasive nature of tracking technologies necessitates a deep dive into their ethical and societal implications. Future research must aim to balance innovation with ethical considerations, developing guidelines that ensure the responsible deployment and utilization of tracking systems.

\section{Conclusion}
\label{sec:conclusion}
We examined the integration of DOT and TP, highlighting their transformative potential across various sectors. Our critical analysis of current methodologies and exploration of emerging technologies has unveiled significant advancements and outlined persistent challenges. The synergistic analysis between DOT and TP highlights the enhanced precision and operational efficiency achievable, opening new avenues for advanced tracking system development.

This study sets a new benchmark for future and ethical advancements. The evaluation of existing methodologies provides a critical understanding of the state of the art, aimed at enhancing adaptability, reducing resource intensity, and prioritizing data privacy.

The insights gained from this review are poised to stimulate further academic research and practical applications in developing next-generation DOT and TP systems. Future research should focus on improving the scalability of these systems, their functionality in diverse environments, and the integration of ethical considerations into the design process. This comprehensive analysis serves as a basis for future investigations, pushing the boundaries of current technology towards more capable and efficient tracking systems.
\subsection*{Declarations} 
\bmhead{Acknowledgments}
The authors acknowledge the support of ML-Labs – SFI Centre for Research Training in Machine Learning for funding and support and SFI Insight Centre for Data Analytics.

\bmhead{Author contributions}
Conceptualization, Methodology and writing—original draft preparation, Zhongping Dong;
Wring, Review and Editing, Liming Chen; Validation, Investigation; Wring, Review and Editing, Mohand Tahar Kechadi. All authors have reviewed and approved the final manuscript. 

\bmhead{Funding}
This research work is funded by ML-Labs – SFI Centre for Research Training in Machine Learning (grant
number 18/CRT/6183).

\bmhead{Availability of data and materials}
Data supporting the findings of this study are publicly accessible.

\bmhead{Competing Interests}
The authors of this manuscript are not registered in the editorial system of this journal as editors or
reviewers.

\backmatter
\subsection*{Abbreviations} 
\noindent 
\begin{tabularx}{\textwidth}{@{} >{\bfseries}l X @{}} 
ADE & Average Displacement Error: Measures the average Euclidean distance between predicted and actual positions over all time steps, with lower values indicating higher prediction accuracy. \\
DNN & Deep Neural Network: A neural network with multiple layers, enabling complex feature extraction and pattern recognition. \\
DOT & Dynamic Object Tracking: The real-time process of tracking objects in motion within a given environment. \\
ECE & Expected Calibration Error: Difference between predicted confidence and actual accuracy in object tracking. \\
EKF & Extended Kalman Filter: A recursive filter used to estimate the state of a system in motion, commonly applied in tracking. \\
FDE & Final Displacement Error: Evaluates the Euclidean distance between predicted and actual positions at the final time step, assessing long-term prediction accuracy. \\
GNN & Graph Neural Network: A type of neural network that works with graph structures, effective in capturing dependencies in trajectory prediction. \\
IoU & Intersection over Union: Measures overlap between predicted and ground truth bounding boxes. \\
LiDAR & Light Detection and Ranging: A remote sensing method that uses laser to measure distances, often used for spatial accuracy in tracking. \\
MOT & Multiple Object Tracking: Tracking multiple objects across frames. \\
MOTA & Multiple Object Tracking Accuracy: Combines false positives, false negatives, and identity switches to evaluate tracking performance. \\
MOTP & Multiple Object Tracking Precision: Assesses alignment between predicted and actual object positions. \\
SOT & Single Object Tracking: Tracking a single object across frames. \\
THOR & A high-precision tracking solution integrated into multimodal systems for enhanced contextual data analysis. \\
TP & Trajectory Prediction: Refers to forecasting future object positions based on historical data. \\
UAV & Unmanned Aerial Vehicle: Also known as drones, utilized in tracking and surveillance applications. \\
\end{tabularx}

\bibliography{sn-bibliography}

\end{document}